\documentclass{article}

\usepackage[preprint]{neurips_2022}

\usepackage[utf8]{inputenc} 
\usepackage[T1]{fontenc}    
\usepackage{hyperref}       
\usepackage{url}            
\usepackage{booktabs}       
\usepackage{amsfonts}       
\usepackage{nicefrac}       
\usepackage{microtype}      
\usepackage{xcolor}         

\usepackage{graphicx}
\usepackage{xcolor}
\usepackage{algorithm} 
\usepackage{algpseudocode} 
\usepackage{multirow}
\usepackage{mathtools}
\usepackage{graphicx}
\usepackage{multicol}
\usepackage{tabularx}
\usepackage{adjustbox}
\usepackage{amsmath}
\usepackage{siunitx}
\usepackage{bbding}

\algnewcommand\algorithmicforeach{\textbf{for each}}
\algdef{S}[FOR]{ForEach}[1]{\algorithmicforeach\ #1\ \algorithmicdo}

\title{ZeroBERTo: Leveraging Zero-Shot Text Classification by Topic Modeling}

 \author{ 
Alexandre Alcoforado$^{1}$, Thomas Palmeira Ferraz$^{1,2}$, Rodrigo Gerber$^{1}$, Enzo Bustos$^{1}$, André
\\ \textbf{Seidel Oliveira$^{1}$, Bruno Miguel Veloso$^{3}$, Fabio Levy Siqueira$^{1}$, and Anna Helena Reali Costa$^{1}$}\\
$^{1}$Escola Politécnica, Universidade de São Paulo (USP), São Paulo, Brazil \\
\texttt{\{alexandre.alcoforado,rodrigo.gerber,enzobustos,} \\ \texttt{andre.seidel,levy.siqueira,anna.reali\}@usp.br} \\
$^{2}$Télécom Paris, Institut Polytechnique de Paris, Palaiseau, France \\
\texttt{thomas.palmeira@telecom-paris.fr}\\
$^{3}$Universidade Portucalense \& INESC TEC, Porto, Portugal\\
\texttt{bruno.m.veloso@inesctec.pt }\\
}

\begin{document}

\maketitle

\begin{abstract}
  Traditional text classification approaches often require a good amount of labeled data, which is difficult to obtain, especially in restricted domains or less widespread languages. This lack of labeled data has led to the rise of low-resource methods, that assume low data availability in natural language processing. Among them, zero-shot learning stands out, which consists of learning a classifier without any previously labeled data. The best results reported with this approach use language models such as Transformers, but fall into two problems: high execution time and inability to handle long texts as input. This paper proposes a new model, \texttt{ZeroBERTo}, which leverages an unsupervised clustering step to obtain a compressed data representation before the classification task. We show that \texttt{ZeroBERTo} has better performance for long inputs and shorter execution time, outperforming XLM-R by about $12\%$ in the F1 score in the FolhaUOL dataset.
  

\end{abstract}

\section{Introduction}

The current success of supervised learning techniques in real-world Natural Language Processing (NLP) applications is undeniable.
While these techniques require a good set of labeled data, large corpora of annotated texts are difficult to obtain, as people (sometimes experts) are needed to create manual annotations or revise and correct predefined labels.
This problem is even more critical in languages other than English: statistics\footnote{Statistics available at: \url{https://w3techs.com/technologies/overview/content\_language}.} show that English is used by \SI{63.1}{\percent} of the population on the internet, while Portuguese, for instance, is only used by \SI{0.7}{\percent}. This scenario has contributed to the rise of Low-Resource NLP, which aims to develop techniques to deal with low data availability in a specific language or application domain \citep{hedderich2020survey}. 

Recently, the concept of \textit{zero-shot learning} emerged in NLP: a semi-supervised approach in which models can present results equivalent to those of supervised tasks, such as classification in the absence of labeled data.
Current approaches to the zero-shot text classification task ({\textsc{0shot-TC}}) make use of the good performance that \textit{Transformers} have demonstrated in text entailment tasks \citep{yin2019benchmarking}. 
In order to be able to process text in a way that is not uniquely suited to any specific task or data-set, these Transformers are first pre-trained in large general databases (usually taken from Wikipedia) and then fine-tuned into a small mainstream data-set for the natural language inference task (such as GLUE \citep{wang2019glue} and XNLI \citep{conneau2018xnli}). 
However, the use of models based entirely on Transformers falls into two critical problems: (\textit{i}) limitation of the maximum size of the input text, and (\textit{ii}) long run-time for large volumes of data.
While there are transformer-based solutions to these problems individually \citep{beltagy2020longformer,sanh2019distilbert,Zaheer2020BigBT}, to the best of our knowledge, there is no solution that addresses both, nor even in the context of \textsc{0shot-TC}. 

In this paper, we propose a new hybrid model that merges Transformers with unsupervised learning, called \texttt{ZeroBERTo} -- Zero-shot BERT based on Topic Modeling --, which is able to classify texts by learning only from unlabeled data. 
Our contribution not only handles long inputs -- not limiting the input size and considering every input token to encode the data -- but also offers a faster execution time.
We propose an experimental setup with unlabeled data, simulating low-resource scenarios where real-life NLP researchers may find themselves. Then, we compare \texttt{ZeroBERTo} to a fully-Transformer-based zero-shot on a categorization dataset in Portuguese, FolhaUOL\footnote{Available at: \url{https://www.kaggle.com/marlesson/news-of-the-site-folhauol}.}.
Our results show that our model outperforms the previous one, in the best scenario, with about \SI{12}{\percent} better \textit{label aware weighted F1-score} and around 13 times faster total time.

The paper is structured as follows: Sect. \ref{sec:background} presents a background of how it is possible to move from data scarcity to zero-shot learning, as well as the related work on getting the best model for the \textsc{0shot-TC} task. Sect. \ref{sec:proposedframework} formalizes the \texttt{ZeroBERTo} task and describe its training and inference procedures. Then, Sect. \ref{sec:experiments} describes the experimental setup that makes it possible to simulate low-resource scenarios to evaluate the proposed model. Finally, the discussion of the results of the experiments along with our final remarks is in Sect. \ref{sec:discussion}.

\section{Background and Related Work}
\label{sec:background}
The first approach to overcome the shortage of labeled data for classification suggests adopting 
data augmentation strategies \citep{jacobs1992joining}, relying on methods to generalize from small sets of already annotated data. Still, the problem persists when there is an extreme lack of data. 
An alternative approach is to treat the task as a topic modeling problem. Topic modeling is an unsupervised learning technique capable of scanning a set of documents, detecting patterns of words and phrases within them, and automatically clustering groups of similar words and expressions that best characterize a set of documents \citep{chen2016short}.
There is usually a later labeling step for these clusters, which can be a problem as their interpretation is sometimes challenging, and a labeling error can affect the entire process. An automatic method for this is in our interest.

The context presented helps explain the growing interest in the field of Low-Resource NLP \citep{chang2008importance,hedderich2020survey}, which addresses traditional NLP tasks with the assumption of scarcity of available data. Some approaches to this family of tasks propose semi-supervised methods, such as adding large quantities of unlabeled data to a small labelled dataset \citep{nigam2000text}, or applying cross-lingual annotation transfer learning \citep{cross-lingualtransfer} to leverage annotated data available in languages other than the desired one. Other approaches try to eliminate the need for annotated data for training, relying, for example, on pre-trained \textit{task-agnostic} neural language models \citep{meng2020text}, which may be used as language information sources, as well as representation learning models \citep{ji2014representation} for word, sentence or document classification tasks.

Recent breakthroughs in pre-trained neural models have expanded the limits of what can be done with data shortage. The Transformer model \citep{attentionallyouneed}, which relies solely on attention mechanisms for learning, followed by BERT \citep{BERT} -- a pre-trained Transformer encoder capable of deep bidirectional encoding -- offered the possibility of using general-purpose models with previous language understanding. With little to no fine-tuning, BERT-like models have been successfully applied to most natural language comprehension tasks \citep{fewshot}, and also show a significant reduction in the need for training data \citep{brown2020language}. Such models tend to work better for the  \textsc{0shot-TC} task, as they carry information about the context and semantic attributes within their pre-trained parameters. On the downside, pre-trained Transformers are complex, with millions of trainable parameters and slow processing of large quantities of data, and due to memory issues, most pre-trained Transformers cannot process inputs larger than 512 tokens at a time. Also, attention models have another problem related to input size: attention cannot keep track of all information present in a large text, worsening the performance of the models.

In this context, zero-shot learning approaches stand out \citep{socher2013zero}. A simple way to explain zero-shot is to compare its paradigm with humans' ability to recognize classes of objects by having a high-level description of them without seeing an example of the object previously. \citet{yin2019benchmarking} defines that \textsc{0shot-TC} aims to learn a classifier $f: X \rightarrow Y$, whereas classifier $f(.)$, however, does not have access to data $X$ specifically labeled with classes from $Y$.
We can use the knowledge that the model already has to learn an intermediate layer of semantic attributes, which is then applied at inference time to recognize unseen classes during the training stages \citep{zhang2019integrating}. 

Several works that seek to improve the performance of zero-shot learning inspired ours. \citet{li2015semi} worked in the image domain, seeking to develop a two-stage model that first learns to extract relevant attributes from images automatically and then learns to assign these attributes to labels. Our proposal performs the same two steps for the text classification problem but does not use any specific knowledge of external data or require any labelled data. 

In the text domain, \citet{mekala2020contextualized} defines weak-supervised learning similar to our definition of zero-shot learning. With unlabeled data and a list of classes as inputs, it applies seed word lists to guide an interactive clustering preview. \citet{meng2020text} uses topic mining to find out which words have the same semantic meaning as the proposed labels, and with that makes a fine-tuning of the language model assuming the implicit category as the presence of these words in the text. Unlike these approaches, our model does not require the user to have any seed word for the labels, and instead of automatically learning them from the labels themselves, \texttt{ZeroBERTo} discovers them from the input data through topic modeling and then assigns them to the labels based on the language model used.

\section{Proposed Method}
\label{sec:proposedframework}

In this section, we introduce \texttt{ZeroBERTo} which leverages Topic Modeling and pre-trained Language Models (LMs) for the task of zero-shot multi-class text classification (\textsc{0shot-TC}). 

\subsection{\textsc{0shot-TC} Task Formalization}

Given a set of unlabeled documents $\mathcal{D} = \mathcal\{d_{1}, d_{2}, \ldots, d_{n}\}$ and a set of \textit{m} semantically disjoint and relevant label names $\mathcal{L} = \mathcal\{l_{1}, l_{2}, \ldots, l_m \}$, \textsc{0shot-TC} aims to learn 
$f: \mathcal{D} \times \mathcal{L} \rightarrow \Theta$, $|\Theta| = |\mathcal{L}|$ and $\Theta$ defines a probability $\theta_{j}^{i} \in [0,1]$ for each label ${l}_{j}$ being the label for $d_{i}$ \citep{yin2019benchmarking}. A single-label classification of a document $d_{i}$ may then be carried out as $l_j \in \mathcal{L} \  | \  j = argmax_{(j)} (\theta_{1}^{i}, \theta_{2}^{i}, \ldots, \theta_{m}^{i})$ -- as a notation simplification, for now on, we mention this as $argmax_{(l \in \mathcal{L})} ({\Theta}_{i})$.

Standard approaches to the \textsc{0shot-TC} task treat it as a \textit{Recognizing Textual Entailment (RTE)} problem: given two documents $d_{1}$, $d_{2}$, we say ``$d_{1}$ \textit{entails} $d_{2}$" ($d_{1} \Rightarrow d_{2}$) if a human reading $d_{1}$ would be justified in inferring the proposition expressed by $d_{2}$ (named \textit{hypothesis}) from the proposition expressed by $d_{1}$ \citep{korman2018defining}. 
In the case of \textsc{0shot-TC}, $d_{2}$ is the hypothesis $\mathcal{H}(l_{j})$, which is simply a sentence that expresses an association to $l_{j}$. For example, in a news categorization problem, a label could be ``\textbf{sports}" and a hypothesis for it could be ``This news is about \textbf{sports}". Creating the hypothesis is essential to make it understandable by a Language Model, and allows us to discover the probability $P(l_{j} | d_{i})$ = $P(d_{i} \Rightarrow \mathcal{H}(l_{j}))$, as $P(d_{i} \Rightarrow \mathcal{H}(l_{j}))$ can easily be inferred by a LM, using $d_{i}$ and $\mathcal{H}(l_{j})$ as inputs. For the zero-shot text classification task, it calculates the textual entailment probability of each possible label. This inference, however, is quite demanding computationally.
 
\subsection{\texttt{\bf ZeroBERTo}}

\texttt{ZeroBERTo} works differently: instead of processing the entire document in the LM, it learns a compressed data representation in an unsupervised way and only processes this representation in the LM. Thus, it is possible to obtain better performance with large inputs and shorter total time than the standard model, even considering the training time added by the unsupervised step.

To learn this representation, \texttt{ZeroBERTo} uses a statistical model, named \textbf{Topic Modeling} (TM), which examines documents and discovers, from the statistics of the words that occur in them, which abstract ``topics'' are covered, discovering hidden semantic structures in a text body.
Given a set of unlabeled documents $\mathcal{D}$, TM aims at learning a set of topics $\mathcal{T}$. 
A topic ${t} \in \mathcal{T}$ is a list of $q$ words or \textit{n-grams} that are characteristic of a cluster but not of the entire documents set. Then, TM also learns how to represent any document ${d}_{i} \in \mathcal{D}$ as a composition of topics expressed by $\Omega_{TM}({d}_{i}) = (\omega_{1}^{i}, \omega_{2}^{i}, \ldots, \omega_{k}^{i})$, such that $\omega_{k}^{i}$ denotes the probability of a document ${d}_{i}$ belonging to a topic ${t}_{k}$. 

With this in place, instead of analyzing the relation between document ${d}_{i}$ and label ${l}_{j}$, we determine the entailment between the learned topic representation $\Omega_{TM}({d}_{i})$ of each document and each label ${l}_{j}$. Topics found are given as input to the LM, as a text list of words/expressions that represent the topic, in order to infer entailment probabilities. If the topic representation was learnt properly, then we can assume independence between $l_{j}$ and $d_{i}$ given a topic $t_{k}$, therefore $P(l_{j} | t_{k}, d_{i}) = P(l_{j} | t_{k}) = P(t_{k} \Rightarrow \mathcal{H}(l_{j}))$.
We then solve the \textsc{0shot-TC} task by calculating the compound conditional probability
\begin{equation}   \mathcal{\theta}_{i}^{j} =  P(l_{j} | d_{i}) =  \sum_{t_k\in \mathcal{T}}{P(l_{j} | t_{k}) * P(t_{k} | d_{i})} = \sum_{t_k\in \mathcal{T}}{P(t_{k} \Rightarrow \mathcal{H}(l_{j})) * \Omega_{TM}^{k}({d}_{i})} 
\label{eq:equation}
\end{equation}
for each label ${l}_{j}$ to determine  ${\Theta}_{i}$ = $(\mathcal{\theta}_{i}^{1},\mathcal{\theta}_{i}^{2}, \ldots, \mathcal{\theta}_{i}^{m})$. Classification is then carried out by selecting $argmax_{(l \in \mathcal{L})} ({\Theta}_{i}) $.  

\textbf{Algorithm 1}: Given a set of documents $\mathcal{D}$, a set of labels $\mathcal{L}$, a hypotesis template $\mathcal{H}$, a topic model $TM$ and a Language model $LM$ as input, \linebreak \texttt{ZeroBERTo-training} (see Alg. \ref{alg:zerobertotrain}) returns a trained model $\mathcal{Z}$. For that, it trains $TM$ on $\mathcal{D}$ using $TM$\textsc{.fit} (line 2), that learns the topic representation of those documents. Then, it applies \textsc{LM.predict} for all topics learned in $TM$ (lines 4 to 7). This function, given a topic $t_{k}$, returns the set of probabilities $P(t_{k} \Rightarrow \mathcal{H}(l_{j}))$ for all $l_{j} \in \mathcal{L}$. In the end, the model $\mathcal{Z}$ gathers all information learned from $\mathcal{D}$. 

\textbf{Algorithm 2}: \texttt{ZeroBERTo-prediction} leverages a trained model $\mathcal{Z}$ and a specific document $d_{i}$ to return the predicted label $l \in \mathcal{L}$ (see Alg. \ref{alg:zerobertopredict}). For this, it uses $\mathcal{Z}.TM.$\textsc{TopicEncoder} (line 1), that returns the topic representation $\Omega_{TM}({d}_{i})$ of $d_{i}$. This was learned by $\mathcal{Z}.TM$ in Alg. \ref{alg:zerobertotrain}. Then, it calculates the equation (\ref{eq:equation}) for all candidate labels (lines 2 to 8), returning the one with maximum probability.

\begin{multicols}{2}
    \begin{algorithm}[H]
        \footnotesize
    	\caption{ZeroBERTo-training} 
    	\label{alg:zerobertotrain}
    	\begin{algorithmic}[1]
    	\Require ${{\mathcal{D}}, \mathcal{L}, \mathcal{H}, TM, LM}$
    	\Ensure $\mathcal{Z}$
 
            \State \textbf{create} $\mathcal{Z}$ \Comment{Instantiate ZeroBERTo}
        	\State $TM.\textsc{fit}({\mathcal{D}})$ \Comment{Topic Model Training}
        	\State $\mathcal P \leftarrow \{\}$
        	\ForEach {${t_{k}}  \in TM.topics $}
        	    \State $p_{k} \leftarrow LM.predict({t_{k}},\mathcal{H}(\mathcal{L}))$ 
        	    \State $\mathcal P \leftarrow \mathcal P \cup \{ p_{k}\}$ 
        	\EndFor
        	\State $\mathcal{Z}.TM \leftarrow TM $
        	\State $\mathcal{Z}.\mathcal{P},\mathcal{Z}.\mathcal{L} \leftarrow \mathcal{P}, \mathcal{L} $
            \State \Return $\mathcal{Z}$
        
    	\end{algorithmic}
    \end{algorithm}

 \columnbreak

\begin{algorithm}[H]
    \footnotesize
	\caption{ZeroBERTo-prediction} 
	\label{alg:zerobertopredict}
	\begin{algorithmic}[1]
	\Require ${\mathcal Z, d_{i}}$
	\Ensure $l$
    	\State $\Omega_{\mathcal{TM}}^{i} \leftarrow \mathcal{ Z}.TM.\textsc{TopicEncoder}(d_{i}$)
    	
        \State $\Theta_{i} \leftarrow \{\}$
    	        \ForEach {$l_{j} \in \mathcal{Z.L}$}
    	           \State $\theta_{j}^{i} \leftarrow 0 $ 
    	            \ForEach {${t_{k}} \in \mathcal Z.TM.topics $}
            	    \State $ \theta_{j}^{i} \leftarrow \theta_{j}^{i} + ( \mathcal P(t_{k}) * \Omega_{TM}^{i}({t}_{k}))$
            
        	    \EndFor
	    	    \State $\Theta_{i} \leftarrow \Theta_{i} \cup \{\theta_{j}^{i}\}$ 
        	\EndFor
        	\State \Return $argmax_{(l \in \mathcal{L})} ({\Theta}_{i}) $
	\end{algorithmic}
\end{algorithm}
\end{multicols}

\section{Experiments}
\label{sec:experiments}
In this section, we present the experiments performed to validate the effectiveness of \texttt{ZeroBERTo}. Considering that it would be difficult to evaluate our model in a real low-resource scenario, we propose an experimental setup to simulate low-resource situations in labeled datasets. We compare \texttt{ZeroBERTo} with the XLM-R Transformer, fine-tuned only on the textual entailment task. To perform the unsupervised training and evaluation, we use FolhaUOL dataset\footnote{Available at: \url{https://www.kaggle.com/marlesson/news-of-the-site-folhauol}.}.
\subsection{Dataset} 
The FolhaUOL dataset is from the Brazilian newspaper ``Folha de São Paulo'' and consists of 167,053 news items labeled into journal sections (categories) from January 2015 to September 2017. Categories too broad, that do not have a semantic meaning associated with a specific context (as the case of ``editorial" and ``opinion"), were removed from the dataset keeping only the categories presented in Table \ref{tab:FolhaUOL}. For each news article, we take the concatenation of its title and content as input. Table \ref{tab:FolhaUOL} presents the data distribution by category.

\begin{table}[]
\caption{Number of articles by news category within FolhaUOL dataset after cleaning and organizing the data.}
\label{tab:FolhaUOL}

\small
\centering 
\begin{adjustbox}{width=1\textwidth,center}
\begin{tabular}{|c|c||c|c|}
\hline
\textbf{Category} & \textbf{\# of articles} & \textbf{Category}   & \textbf{\# of articles} \\ \hline
Poder e Política            & 22022 & Educação & 2118 \\ \hline
Mercado    & 20970 & Turismo   & 1903 \\ \hline
Esporte              & 19730 & Ciência   & 1335 \\ \hline
Notícias dos Países       & 17130 & Equilíbrio e Saúde    & 1312 \\ \hline
Tecnologia          & 2260  & Comida      & 828  \\ \hline
TV, Televisão e Entretenimento  & 2123 & Meio Ambiente & 491 \\ \hline
\end{tabular}%
\end{adjustbox}

\end{table}

\subsection{Models} We compare our proposal to the XLM-R model. 
\begin{description}
\item \textbf{XLM-R} is the short name for XLM-RoBERTa-large-XNLI, available on Hugging Face\footnote{Available at: \url{https://huggingface.co/joeddav/xlm-roberta-large-xnli}}, which is state of the art in Multilingual \textsc{0shot-TC}. It is built from XLM-RoBERTa \citep{conneau2020unsupervised} pre-trained in 100 different languages (Portuguese among them), and then fine-tuned in the XNLI \citep{conneau2018xnli} and MNLI \citep{williams2018broad} datasets (which do not include the Portuguese language). It is already in the zero-shot learning configuration described by \citet{yin2019benchmarking} with template hypothesis as input. 
The template hypothesis used was ``\emph{O tema principal desta notícia é \{\}}'' and texts larger than the maximum size of XLM-R (512 tokens) are truncated.
\item \textbf{ZeroBERTo} The implementation of our model here makes use of BERTopic \citep{grootendorst2020bertopic} with M-BERT-large (Multilingual BERT) \citep{BERT} as topic modeling step, and the same XLM-R described above as the Transformer for associating the topic representation of each document to labels. Repeating the use of XLM-R seeks to make the comparison fair. BERTopic's hyperparameters are: interval n for n-grams to be considered in the topic representation (\texttt{n\_grams\_range} $\in \{1, 2, 3\}$); number of representative words/n-grams per topic (\texttt{top\_n\_words} = 20); and minimum amount of data in each valid topic (\texttt{min\_topic\_size} = 10). The XLM-R template hypothesis used is ``\emph{O tema principal desta lista de palavras é \{\}}''.
\end{description}

\subsection{Evaluation}

To simulate real-world scenarios, we propose a variation of stratified $k$-fold cross-validation \citep{refaeilzadeh2009cross}. First, we split the data into $k$ disjoint stratified folds, \textit{i.e.} the data were evenly distributed in such a way as to make the distribution of the classes in the $k$ folds follow the distribution in the entire dataset. Next, we use these $k$-folds to perform the following 4 experiment setups:
\begin{description}
\item[Exp. 1 - Labeling a dataset:] Simulates a situation where one needs to obtain the first labeling of a dataset. \texttt{ZeroBERTo} is trained in $(k-1)$ folds and has the performance compared to XLM-R in the same $(k-1)$ folds, in order to assess its ability to label data already seen. Since this is unsupervised learning, evaluating the model's labeling ability in the training data makes sense as it was not exposed to the ground truth labels.
\item[Exp. 2 - Building a model for additional inferences:] Simulates a situation where the researcher wants to create a current model in a real-life application without having data labeled for it. \texttt{ZeroBERTo} is trained in $(k-1)$ folds and can infer new data compared to XLM-R on the remaining fold.
\item[Exp. 3 - Labeling a smaller dataset:] Simulates situation of scarcity of data in which, besides not having labeled data, little data is present. \texttt{ZeroBERTo} is trained in one fold and compared to XLM-R in the same fold. Considering the topic-representation learning stage, the presence of little data could be a bottleneck for \texttt{ZeroBERTo} since the topic representation may not be properly learned.
\item \textbf{Exp. 4 - Building model for additional inferences but with a scarcity of training data:} simulates again how the model would behave in a real-life application with few training data. \texttt{ZeroBERTo} is trained in $1$ fold and compared to XLM-R in the remaining $k-1$ folds. 
\end{description}
We evaluated the performance of both models for each experiment with the following label-aware metrics: weighted-average Precision (P), weighted-average Recall (R), and weighted-average F1-score (F1). For the $k$-fold CV, we use $k = 5$. Experiments were run on an Intel Xeon E5-2686 v4 2.3GHz 61 GiB CPU and an NVIDIA Tesla K80 12 GiB GPU using the PyTorch framework. To run XLM-R, we use batches sized 20 to prevent GPU memory overflow.
\begin{table}[]
\caption{Table shows the results of the experiments for the FolhaUOL dataset. P is weighted-average Precision, R is weighted-average Recall, and F1 is weighted-average F1-score.}
\label{tab:resultsFolhaUOL}
\begin{adjustbox}{width=1.0\textwidth,center}
\begin{tabular}{c|c|c|c|c|c|c|c|c|}
\cline{2-9}
 &
  \multicolumn{2}{c|}{\textbf{Exp. 1}} &
  \multicolumn{2}{c|}{\textbf{Exp. 2}} &
  \multicolumn{2}{c|}{\textbf{Exp. 3}} &
  \multicolumn{2}{c|}{\textbf{Exp. 4}} \\ \cline{2-9} 
 &
  \textbf{XLM-R} &
  \textbf{\texttt{ZeroBERTo}} &
  \textbf{XLM-R} &
  \textbf{\texttt{ZeroBERTo}} &
  \textbf{XLM-R} &
  \textbf{\texttt{ZeroBERTo}} &
  \textbf{XLM-R} &
  \textbf{\texttt{ZeroBERTo}} \\ \hline
\multicolumn{1}{|c|}{\textbf{P}} &
  $0.47 \pm 0.00$ &
  $\mathbf{0.66 \pm 0.01}$ &
  $\mathbf{0.46 \pm 0.01}$ &
  $0.16 \pm 0.08$ &
  $0.46 \pm 0.01$ &
  $\mathbf{0.64 \pm 0.01}$ &
  $\mathbf{0.47 \pm 0.00}$ &
  $0.29 \pm 0.17$ \\ \hline
\multicolumn{1}{|c|}{\textbf{R}} &
  $0.43 \pm 0.00$ &
  $\mathbf{0.54 \pm 0.01}$ &
  $\mathbf{0.43 \pm 0.00}$ &
  $0.21 \pm 0.05$ &
  $0.43 \pm 0.00$ &
  $\mathbf{0.56 \pm 0.02}$ &
  $\mathbf{0.43 \pm 0.00}$ &
  $0.31 \pm 0.12$ \\ \hline
\multicolumn{1}{|c|}{\textbf{F1}} &
  $0.43 \pm 0.00$ &
  $\mathbf{0.54 \pm 0.01}$ &
  $\mathbf{0.42 \pm 0.01}$ &
  $0.15 \pm 0.07$ &
  $0.42 \pm 0.01$ &
  $\mathbf{0.52 \pm 0.02}$ &
  $\mathbf{0.43 \pm 0.00}$ &
  $0.19 \pm 0.17$ \\ \hline \hline
\multicolumn{1}{|c|}{\textbf{Time}} &
  61h30min &
  \textbf{9h21min} &
  15h22min &
  \textbf{6h20min} &
  15h22min &
  \textbf{1h10min} &
  61h30min &
  \textbf{2h25min} \\ \hline
\end{tabular}%
\end{adjustbox}

\end{table}

\subsection{Results}
\begin{figure}
\centering
\includegraphics[width=1.01\textwidth]{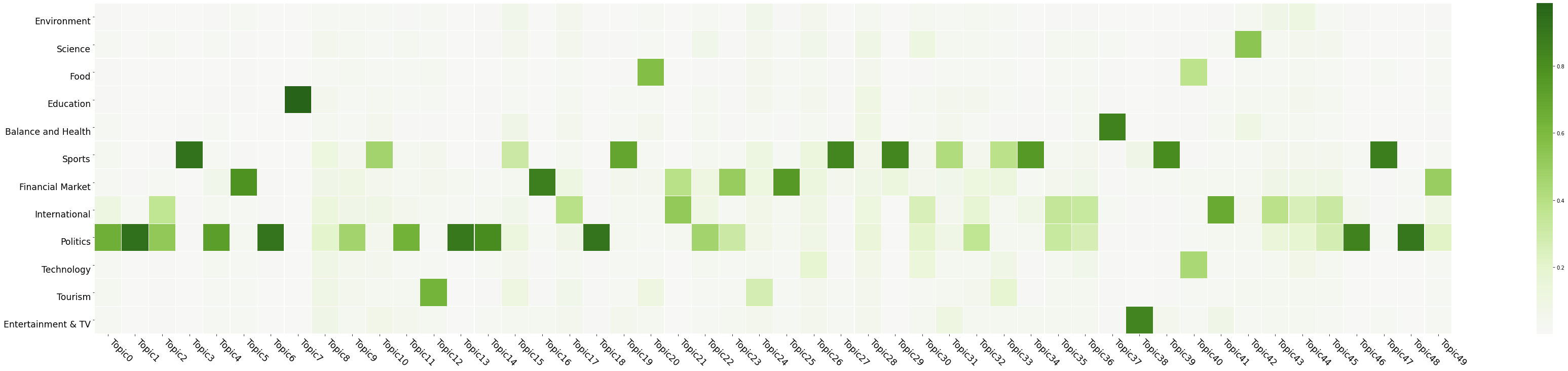}
\caption{Figure shows text entailment results between topics (X-axis) and labels (Y-axis) for the first 50 Topics (sorted by size) in fold 0 from Experiment 3. In total, 213 topics were generated in this experiment.} \label{fig:heatmap}
\end{figure}

Table \ref{tab:resultsFolhaUOL} shows the results of the proposed experiments. Time for \texttt{ZeroBERTo} considers 
unsupervised training time and inference time. Further, as no training is required, only a single run of XLM-R was done on all data. Thus, the times for XLM-R are estimated. Nevertheless, in all experiments, the total time (training + execution) of \texttt{ZeroBERTo} was much lower than the execution time of XLM-R. Our model surpassed XLM-R in all metrics in the experiments in which the evaluation was performed on the data used in the unsupervised training (Exp. 1 and 3). Figure \ref{fig:heatmap} presents a visualization for the entailment mechanism between topics and labels represented by term $P(l_{j} | t_{k}) = P(t_{k} \Rightarrow \mathcal{H}(l_{j}))$ in equation (\ref{eq:equation}). The darker the green, the greater the conditional odds.

\section{Discussion and Future Work}
\label{sec:discussion}
The experiments simulated low-resource scenarios where a zero-shot text classifier can be useful. The results showed that it is possible to obtain a better performance in the \textsc{0shot-TC} task with the addition of an unsupervised learning step that allows a simplified representation of the data, as proposed by \texttt{ZeroBERTo}. Moreover, the proposed model presents itself as an excellent tool to help researchers deal with low-resource scenarios, such as the need to label an entire dataset without any previously labeled training data. 
Another interesting feature is that the model showed evidence of robustness for smaller amounts of data.
In experiment 3, it was trained with \SI{25}{\percent} of the data from experiment 1 and got similar performance metrics in lower time, refuting our concern that little data could be a bottleneck in the model.

However, for configurations where \texttt{ZeroBERTo} was tested simulating real-life applications (Exp. 2 and 4), being exposed to new data, the performance was worse than XLM-R. The results suggest it occurs due to the inability of the embedded topic model to adequately represent new data as a composite of previously learned topics, overfitting training data. This is clear from observing the high variance of the metrics among the k-folds. It allows us to conclude that, for now, the scenarios presented in experiments 1 and 3 are more suitable for using the model.

We have 0.54 of F1-score in the best scenario regarding the metrics obtained. Despite being a positive result considering that it is a multi-class classification, there is much room for improvement. The main reason to be pointed out for low performances is the use of multilingual models that were not fine-tuned in the Portuguese language, which is quite impressive. 

A critical remark to be made is concerning the memory and time trade-off. For example, \texttt{ZeroBERTo} was more than 10x faster than XLM-R in Exp. 3. However, the topic model used by \texttt{ZeroBERTo} bases its clustering on the HDBSCAN method, which reduces time taken for data processing but increases the need for memory \citep{mcinnes2017accelerated}, which XLM-R does not do. As the size of input data grows, processing may become unfeasible. XLM-R, on the other hand, does not use any interaction between data and can be processed in parallel and distributed without any negative effect on the final result. It should be noted, however, that \texttt{ZeroBERTo} does not depend on BERTopic and can use other Topic Modeling techniques that address this issue more adequately in other scenarios.

A significant difficulty of this work was that, as far as the authors are aware of, there are no large benchmark datasets for multi-class text classification in Portuguese, nor general use datasets with semantically meaningful labels. In this sense, some future work directions involve the production of benchmark datasets for Portuguese text classification (and \textsc{0shot-TC}). It would also be interesting to produce Natural Language Inference datasets in Portuguese, which could, in addition to the existing ones \citep{fonseca2016assin,real2020assin}, enable fine-tuning of Transformers \SI{100}{\percent} in Portuguese. Then, it would be possible to compare the performance of the models using BERTimbau (BERT-Portuguese) \citep{souza2020bertimbau} both in clustering and classifying. It would also be worthwhile to test the proposed model in other domains: to name one, legislative data present similar challenges \citep{ferraz2021debacer}. Another interesting future work would be to enable \texttt{ZeroBERTo} to deal with multi-label classification, where each document can have none, one or several labels.

\section*{Acknowledgments}
This research was supported in part by \textit{Ita\'{u} Unibanco S.A.}, with the scholarship program of \textit{Programa de Bolsas Ita\'{u}} (PBI), and by the Coordenação de Aperfeiçoamento de Pessoal de Nível Superior (CAPES), Finance Code 001, CNPQ (grant 310085/2020-9), and USP-IBM-FAPESP \textit{Center for Artificial Intelligence} (FAPESP grant 2019/07665-4), Brazil.
Any opinions, findings, and conclusions expressed in this manuscript are those of the authors and do not necessarily reflect the views, official policy, or position of the financiers.

\bibliographystyle{plainnat}
\bibliography{mybibliography}

\end{document}